# Light-weight place recognition and loop detection using road markings

Oleksandr Bailo, Francois Rameau, In So Kweon
Department of Electrical Engineering, KAIST, Daejeon, 34141, Korea
(Tel : +82-10-4943-1404; E-mail: {obailo, frameau}@rcv.kaist.ac.kr, iskweon77@kaist.ac.kr)

*Abstract*—In this paper, we propose an efficient algorithm for robust place recognition and loop detection using camera information only. Our pipeline purely relies on spatial localization and semantic information of road markings. The creation of the database of road markings sequences is performed online, which makes the method applicable for real-time loop closure for visual SLAM techniques. Furthermore, our algorithm is robust to various weather conditions, occlusions from vehicles, and shadows. We have performed an extensive number of experiments which highlight the effectiveness and scalability of the proposed method.

## 1. Introduction

Visual Simultaneous Localization and Mapping (SLAM) entirely relies on low-level image features to jointly estimate the structure and the motion of the cameras. This process is prone to errors due to many factors like moving objects, lighting conditions, etc. This becomes a significant problem as the errors accumulate over time which could potentially result in a biased outcome. To limit this effect, loop closure is a common technique to correct the drift. Therefore, being able to robustly detect already visited places in real-time is an essential component in any recent SLAM algorithm. Existing approaches can be divided into two categories: hand-crafted and learnt features. Regarding the first one, López *et al.* [1] have proposed a method for visual place recognition using bag of words computed from binary image features. Alternatively, other works rely on deep learning tools to obtain robust features to represent the image. For instance, Zhou *et al.* [2] have proposed a new approach based on CNN to obtain deep features for scene recognition. More recently, NetVLAD [3] have been developed to learn image representation in a weakly supervised manner. In term of place recognition accuracy, methods relying on CNN perform better than the ones that utilize hand-crafted features. However, they tend to be comparably slower and require expensive computational resources (*e.g.* GPUs).

On the other hand, several works have attempted to utilize robust landmarks like road landmarks (*i.e.* markings or signs) for precise localization to compensate the low accuracy of GPS in urban canyons. Indeed, using road markings for place recognition has several advantages. First of all, the matching and storage cost can be significantly reduced compared to traditional feature-based methods. In particular, place recognition can be simply achieved by utilizing road marking labels and their locations, while traditional methods require the comparison of high-dimensional feature spaces. Secondly, road markings are extremely robust to various weather conditions. While visual appearance of the site can change over time, road markings are more robust, static, and precise. Lastly, a combination of various constraints can be enforced in the case of road markings. For instance, the matching can be based on both spatial localization and semantic attributes to improve the efficiency. In the light of these advantages, Wu *et al.* [4] automatically recognize road markings and use them to estimate the precise location of the vehicle. Subsequently, this work has been extended by Ranganathan *et al.* [5] by building a light-weight database using an accurate D-GPS and performing vehicle localization thanks to a lookup in the database. Similarly, Wei *et al.* [6] have proposed to associate the online detection from cameras with a road database to adjust the vehicle pose accordingly. As mentioned previously, most of the landmark-based vehicle localization methods are composed of two phases: database generation, and database lookup. During the database generation stage, visual markings are added to the database thanks to their GPS position estimation. Then, the lookup in the database is performed based on the approximate GPS position of the detected landmarks.

In this work, we do not rely on any external navigation sensors like GPS or IMU and perform database creation as well as place recognition on the fly, purely relying on road markings classification and their relative positions obtained by SLAM. This drastically improves the computational time while maintaining a high-level of accuracy of place recognition even for large-scale scenarios. Moreover, this allows us to achieve real-time loop closure detection for any visual SLAM under urban conditions. To sum up, the contributions of this paper are the following:

- Efficient place recognition and loop detection algorithm based on road markings.
- Extensive number of experiments to demonstrate the robustness and scalability of the proposed method.
- Efficient codes and dataset are made available. [1]

## 2. Methodology

In this work, we aim to perform place recognition using only car onboard camera without relying on external navigation sensors. The key idea of the algorithm is to create sequences of successive road markings of size $k$ (defined by the user) on run and simultaneously perform sequence matching using semantic and relative distance constraints together.

---
[1]https://github.com/BAILOOL/PlaceRecognition-LoopDetection

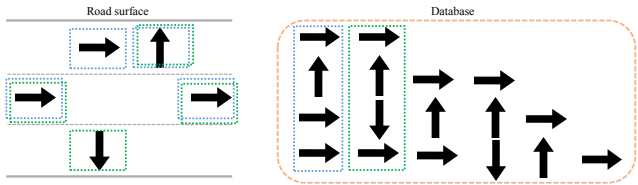

Fig. 1. Database creation: (left) conceptual representation of road marking detection on multiple lanes, (right) overview of the resulting database. For this scenario, two sequences reached the desired size $k = 4$, while remaining incomplete sequences are kept in memory for further updates.

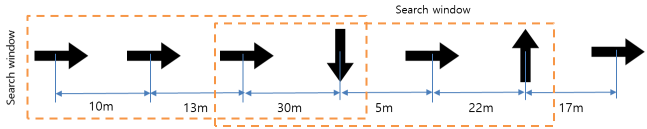

Fig. 2. Example of a sequence in the database with relative distances between entries.

Initially, a road marking detection and recognition algorithm [7] is applied to the current input image to extract road marking bounding boxes as well as their labels. Further, the centroid of every detection is used to estimate the position of the road marking in 3D. Specifically, the position of every road marking is determined using the camera poses in the map (obtained by our own stereo SLAM algorithm) and the camera elevation and orientation relative to the road surface. Then, the road marking 3D position as well as its label are used to update the sequences stored in the database. The sequences of size $k$ are kept unmodified while the incomplete sequences (with less than $k$ elements) are updated (concatenated) with the current road marking. For the sake of efficiency, if several road markings are detected on the same image, they would not update the same sequence in the dataset. The example of the road surface with detected road markings and database overview can be seen in Fig. 1. Due to occlusions and/or misdetections, only several road markings might be detected. Nevertheless, we fully utilize available information by creating all the possible sequences. This approach makes the proposed algorithm robust to the aforementioned cases.

Every time the dataset is updated, the sequence matching function is executed to find candidates. The matching is performed by utilizing both spatial localization and semantic attributes. Specifically, given the search window $k$ (*i.e.* size of the sequences), the algorithm compares every sequence of size $k$ in the dataset by utilizing road markings labels and the relative distance between subsequent road markings (see Fig. 2). If we have $N$ sequences of the length $k$ in the dataset, the time complexity of the algorithm to recognize places is $O(\frac{N(N-1)}{2}k)$.

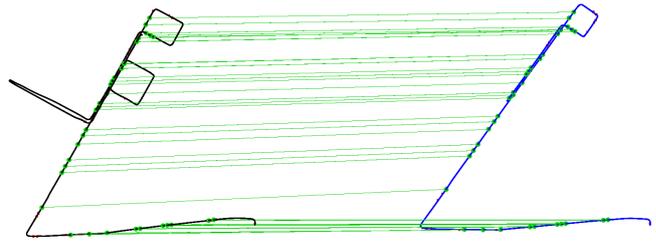

Fig. 3. Matches on large-scale dataset: obtained poses (in black), road markings (in red), matched regions (in green) of the sequence.

## 3. EXPERIMENTS

In order to evaluate the proposed algorithm, we have collected multiple challenging stereo sequences in a crowded urban environment. The first dataset is composed of two long video sequences (about 35min of driving, resulting in 32422 images in total). These two sequences have been acquired with one-week interval and with slightly different camera configuration (baseline, white balance, exposure, etc.) to make this large-scale place recognition even more difficult. The second dataset has been captured specifically to underline the real-time loop detection capability of our approach. This sequence consists of approximately 10000 stereo frames and contains two large loops of 2.5km each. All these data have been acquired via a stereo rig (composed of two PointGrey Flea 3 1280x512p cameras equipped with wide angle aspherical lenses) installed on a car. For all sequences, our SLAM algorithm is applied to obtain the poses of the cameras and the resulting 3D map. At the same time, a road marking detection and recognition module is running to obtain the road marking bounding boxes and labels. Then, the centroids of each road marking as well as their 3D position are estimated.

The first experiment consists in finding the common places between two large-scale datasets acquired at different time and conditions (heavy traffic, different illumination, etc.) where typical place recognition algorithms tend to fail. A sample of the obtained place recognition outputs (with $k = 4$) is provided in Fig. 4(left), while the calculated 3D poses, recognized road markings, and matched places can be seen in the Fig. 3. Qualitatively, it is noticeable that our algorithm retrieves very consistent places locations with a very low number of outliers. Using the second dataset, our goal is to identify loops in a single sequence and perform loop closure to correct the poses and the 3D map. A representative loop detection result is available in Fig. 4(right). In this figure, it is clear that despite a strong illumination variation, our strategy remains very efficient to determine the proper locations to close the loop. The computational time of our algorithm is about 1ms per place recognition inquiry. The computational load is subject to the size of the dataset. However, it systemically remains drastically faster than conventional approaches by multiple order of magnitude.

To better understand the influence of the search window size $k$ on the output of the algorithm we also propose to run the place recognition algorithm with various values of $k$. This experiment has been conducted for both the large-scale place recognition and the online loop detection. The results are visible in the Table 1 and Table 2 respectively. Through this assessment, we can notice a strong correlation between the search window size and the number of matched sequences. For instance, in the loop closure case with a small window's size ($k = 2$), 20 potential candidates are returned since this configuration does not enforce strong constraints, leading to a relatively high number of false detection. However, the percentage of false positive reduces as the search window size increases. For example, in the Table 1, setting the parameter $k = 3$ is enough to return more than $90\%$ of true positive detection.

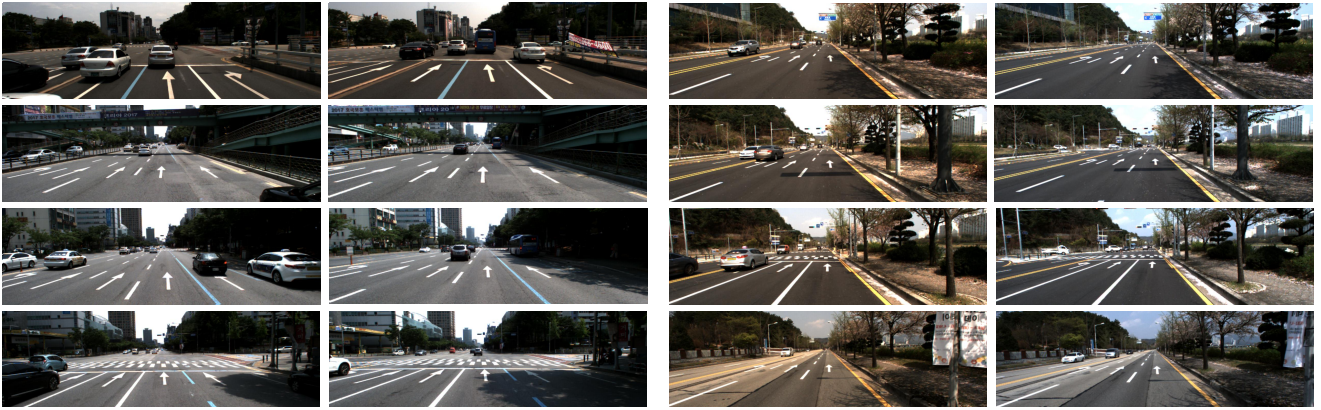

Fig. 4. Representative results of the algorithm ($k = 4$) for: (left) place recognition, (right) loop detection.

TABLE 1
INFLUENCE OF SEARCH WINDOW SIZE ON PLACE RECOGNITION.

| Search window size | 3 | 4 | 5 | 6 | 7 | 8 | 9 | 10 |
|---|---|---|---|---|---|---|---|---|
| Number of loop candidates | 26 | 16 | 12 | 8 | 6 | 4 | 2 | 1 |
| Percentage of correct matches | 92.3 | 100 | 100 | 100 | 100 | 100 | 100 | 100 |
| Percentage of incorrect matches | 7.7 | 0 | 0 | 0 | 0 | 0 | 0 | 0 |

TABLE 2
INFLUENCE OF SEARCH WINDOW SIZE ON LOOP DETECTION.

| Search window size | 2 | 3 | 4 | 5 | 6 |
|---|---|---|---|---|---|
| Number of loop candidates | 20 | 4 | 3 | 2 | 1 |
| Percentage of correct matches | 35 | 100 | 100 | 100 | 100 |
| Percentage of incorrect matches | 65 | 0 | 0 | 0 | 0 |

## 4. CONCLUSION

In this paper, we have proposed an efficient and robust place recognition and loop detection algorithm using road markings. The proposed algorithm is proven to be robust to various weather and environmental conditions. Furthermore, the algorithm creates a database of road markings using images obtained from onboard cameras without the need for external navigation devices. The experimental results confirm the robustness and the scalability of the proposed algorithm. Encouraging results of the proposed method open up new opportunities to explore. Specifically, in the future work, we would like to extend the proposed algorithm to utilize road markings as well as road signs. We believe that combining both road landmarks would further boost the performance and robustness of the algorithm.

ACKNOWLEDGEMENT

This work was supported by the Technology Innovation Program (No. 2017-10069072) funded By the Ministry of Trade, Industry Energy (MOTIE, Korea). The second author was supported by Korea Research Fellowship Program through the National Research Foundation of Korea (NRF) funded by the Ministry of Science, ICT and Future Planning (2015H1D3A1066564).